\documentclass[a4paper,fleqn]{cas-sc}

\usepackage{longtable}
\usepackage[numbers]{natbib}

\def\tsc#1{\csdef{#1}{\textsc{\lowercase{#1}}\xspace}}
\tsc{WGM}
\tsc{QE}

\begin{document}
\let\WriteBookmarks\relax
\def\floatpagepagefraction{1}
\def\textpagefraction{.001}

\shorttitle{Detecting Dementia from Speech and Transcripts using Transformers}    

\shortauthors{L.Ilias, D.Askounis, and J.Psarras}  

\title [mode = title]{Detecting Dementia from Speech and Transcripts using Transformers}  



%

\author[1]{Loukas Ilias}[orcid=0000-0002-4483-4264]

\cormark[1]


\ead{lilias@epu.ntua.gr}



\affiliation[1]{organization={Decision Support Systems Laboratory, School of Electrical and Computer Engineering, National Technical University of Athens},
            addressline={Zografou}, 
            city={Athens},
            postcode={15780}, 
            country={Greece}}

\author[1]{Dimitris Askounis}[orcid=0000-0002-2618-5715
]


\ead{askous@epu.ntua.gr}



\author[1]{John Psarras}[]


\ead{john@epu.ntua.gr}



\cortext[1]{Corresponding author}



\begin{abstract}
Alzheimer's disease (AD) constitutes a neurodegenerative disease with serious consequences to peoples' everyday lives, if it is not diagnosed early since there is no available cure. Alzheimer's is the most common cause of dementia, which constitutes a general term for loss of memory. Due to the fact that dementia affects speech, existing research initiatives focus on detecting dementia from spontaneous speech. However, little work has been done regarding the conversion of speech data to Log-Mel spectrograms and Mel-frequency cepstral coefficients (MFCCs) and the usage of pretrained models. Concurrently, little work has been done in terms of both the usage of transformer networks and the way the two modalities, i.e., speech and transcripts, are combined in a single neural network. To address these limitations, first we represent speech signal as an image and employ several pretrained models, with Vision Transformer (ViT) achieving the highest evaluation results. Secondly, we propose multimodal models. More specifically, our introduced models include Gated Multimodal Unit in order to control the influence of each modality towards the final classification and crossmodal attention so as to capture in an effective way the relationships between the two modalities. Extensive experiments conducted on the ADReSS Challenge dataset demonstrate the effectiveness of the proposed models and their superiority over state-of-the-art approaches.
\end{abstract}



\begin{keywords}
Dementia \sep Speech \sep log-Mel spectrogram \sep Mel-frequency cepstral coefficients \sep Vision Transformer \sep Gated Multimodal Unit \sep Crossmodal Attention
\end{keywords}

\maketitle

\section{Introduction}
Alzheimer's disease (AD) constitutes a neurodegenerative disease, meaning that it becomes worse with time. Moreover, it is thought to begin 20 years or more before symptoms arise, with minor changes in the brain that are unnoticeable to the person affected \cite{alzheimer20192019}. Due to the fact that there is no available treatment, it is important to be diagnosed early before affecting the patient's everyday life to a high degree. However, the diagnosis of Alzheimer's disease through Magnetic Resonance Imaging (MRI), Positron emission tomography (PET) images, or electroencephalogram (EEG) signals has been proven to be time and cost-ineffective. Alzheimer's is the most common cause of dementia, Thus, research works have moved their interest towards dementia identification from spontaneous speech, since the speech of AD patients is affected to a high degree.

Research works using audio data to categorize people into AD and non-AD patients use mainly acoustic features extracted from speech, such as eGeMAPS \cite{7160715}, duration of speech etc. After having extracted the respective feature sets, they train traditional machine learning classifiers, such as Support Vector Machines (SVM), Decision Trees (DT) etc. However, feature extraction constitutes a time-consuming procedure, does not generalize well to data from new patients, and often demands some level of domain expertise. Log-Mel Spectrograms and Mel-frequency cepstral coefficients (MFCCs) are being used extensively in heart sound classification \cite{9175450}, emotion recognition \cite{8817913}, depression detection \cite{zhao20h_interspeech}, etc. In addition, pretrained models on the domain of computer vision, including AlexNet, MnasNet, EfficientNet, VGG, etc., have been exploited extensively in many tasks, including Alzheimer's disease detection through MRIs \cite{8784845}, detection of epileptic seizures using EEG signals \cite{RAGHU2020202, 9392720}, facial emotion recognition \cite{giannopoulos2018deep}, analysis of online political advertisements \cite{sanchez-villegas-etal-2021-analyzing}, heart sound classification \cite{9175450}, voice pathology diagnosis \cite{ghoniem2019deep}, etc. Thus, the representation of speech signal as an image constitutes a motivation for exploiting image-based models. However, limited research has considered speech in such a way \cite{9383491,10.3389/fcomp.2021.624683,10.3389/fcomp.2021.624694}. Therefore, in this study we convert each audio file into an image consisting of three channels, namely log-Mel spectrograms (and MFCCs), their delta, and delta-delta. Contrary to \cite{10.3389/fcomp.2021.624683, 10.3389/fcomp.2021.624694}, we use the delta and delta-delta features for adding more information \cite{5947425,1168654}. Next, we employ many pretrained models, including AlexNet, VGG16, DenseNet, EfficientNet, Vision Transformer, etc. and compare their performances. Our main motivation is to find the best model for extracting acoustic features and exploiting it in the multimodal setting.

Moreover, another limitation of the current research works lies in the usage of multimodal models. To be more precise, research works train first acoustic and language models separately and then use the majority voting approach for classifying people into AD and non-AD patients \cite{cummins20_interspeech,sarawgi20_interspeech,9459113}. This fact increases substantially the training time and does not take into account the inter- and intra-modal interactions. Other research works add or concatenate acoustic and language representations during training \cite{10.3389/fcomp.2021.624683,pan21c_interspeech,10.3389/fnagi.2021.623607}.  This approach may decrease the performance of the multimodal models in comparison with the unimodal ones, since the different modalities are treated equally. In addition, there are studies, which concatenate the features from different modalities at the input level (early fusion approaches) \cite{chen21r_interspeech,pompili20_interspeech,martinc20_interspeech}. Little work has been done in terms of exploiting  techniques to control the influence of each modality towards the final classification and capturing the inter- and intra-modal interactions. Specifically, the authors in \cite{rohanian20_interspeech, rohanian21_interspeech} used feed-forward highway layers with a gating mechanism. However, the authors did not experiment with replacing the gating mechanism with a simple concatenation operation. Thus, the addition of the introduced gating mechanism cannot guarantee increase in the performance. To tackle this limitation, we propose  new methods, which can be trained in an end-to-end trainable way, to combine the representations of the different modalities. Firstly, we convert each audio file into an image consisting of three channels, namely log-Mel spectrograms (and MFCCs), their delta, and delta-delta. We pass these images through a Vision Transformer, which is the best performing model among the proposed pretrained models, i.e., AlexNet, VGG16, DenseNet, EfficientNet, etc. Each transcript is passed through a BERT model. Next, we propose a Gated Multimodal Unit in order to assign more importance to the most relevant modality while suppressing irrelevant information. In addition, we introduce crossmodal attention so as to model crossmodal interactions.

Our main contributions can be summarized as follows:
\begin{itemize}
    \item We propose multimodal deep learning models to detect AD patients from speech and transcripts. We also introduce a multimodal gate mechanism, so as to control the influence of each modality towards the final classification.
    \item We introduce the crossmodal attention and show that crossmodal models outperform the multimodal ones.
\end{itemize}

\section{Related Work}
In this section, we describe the landscape of relevant works on the detection of dementia from spontaneous speech. These works are divided into two approaches, \textit{(i)} methods using only the acoustic modality (Section \ref{only_speech}) and \textit{(ii)} multimodal methods using both the textual (transcripts) and the acoustic modality (Section \ref{multimodal}). Finally, Section \ref{overview} provides an overview of the related work along with the limitations. Also, we discuss how our work differs from the existing research initiatives.

\subsection{Detecting Dementia only from Speech} \label{only_speech}

Meghanani et al. \cite{9383491}  used the ADReSS Challenge Dataset and proposed three deep learning models to detect AD patients using only speech data. Firstly, they converted the audio files into Log-Mel spectrograms and MFCCs along with their delta and delta-delta, in order to create an image consisting of three channels. Next, they divided the images into non-overlapping segments of 224 frames and passed each frame through five convolution layers followed by LSTM layers. In the second proposed model, they replaced the five convolution layers with a pretrained ResNet18 model. Finally, they trained a model consisting of BiLSTMs and CNN layers. Results showed that Log-Mel spectrograms and MFCCs are effective for the AD detection problem. One limitation of this study is that the authors employed only one image-based pretrained model, i.e., ResNet18.

Gauder et al. \cite{gauder21_interspeech}  used the ADReSSo Challenge Dataset and extracted a set of features from speech, namely eGeMAPS \cite{7160715}, trill \cite{shor20_interspeech}, allosaurus \cite{9054362}, and wav2vec2 \cite{baevski2020wav2vec}, where each feature vector was fed into two convolution layers. Then, the outputs of the convolution layers were concatenated and were passed through a global average pooling layer followed by a dense layer, in order to get the final output. Results from an ablation study showed that trill and wav2vec2 constituted the best features. The main limitations of this study are the feature extraction process and the concatenation of the feature representations.

Balagopalan and Novikova \cite{balagopalan21_interspeech}  used the ADReSSo Challenge Dataset and introduced three approaches to differentiate AD from non-AD patients by extracting 168 acoustic features from the speech audio files, computing the embeddings of the audio files using wav2vec2, and finally combining the aforementioned approaches by simply concatenating the two representations. Results showed that a Support Vector Machine trained on the acoustic features yielded the highest precision, whereas the SVM classifier trained on the concatenation of the embeddings achieved the highest accuracy, recall, and F1-score. The limitation of this study lies on the feature extraction process, the train of traditional machine learning classifiers, and the usage of the concatenation operation, where the same importance is assigned to the features.

Ref. \cite{10.3389/fpsyg.2020.623237}  used the ADReSS Challenge Dataset and introduced two approaches targeting at diagnosing dementia only from speech. Firstly, after employing VGGish \cite{7952132}, they  used the features extracted via VGGish and trained shallow machine learning algorithms to detect AD patients. Next, they proposed a convolutional neural network for speech classification, namely DemCNN, and claimed that DemCNN outperformed the other approaches. The main limitation of this research work is the train of shallow machine learning classifiers using the VGGish features, which increase the training time.

The authors in \cite{ammar2019evaluation} proposed a feature extraction approach. Specifically, they extracted 54 acoustic features, including duration, intensity, shimmer, MFCCs, etc. Finally, they trained the LIBSVM with a radial basis kernel function. The limitation of this study lies on the feature extraction process and the train of only one traditional machine learning classifier. In addition, the authors have not applied feature selection or dimensionality reduction techniques.

Research works \cite{10.1145/3175587.3175589,8910399}  used the DementiaBank Dataset and exploited a set of acoustic features along with shallow machine learning classifiers. More specifically in \cite{10.1145/3175587.3175589}, the authors extracted a set of 121 features, including the fundamental frequency, the frequency alteration from cycle to cycle, the F0 amplitude variability, features assessing the voice quality, spectral features, etc. The authors expanded this feature set with some statistical sub-features, i.e., min, max, mean, etc. and thus increased the number of features to 811. After employing feature selection techniques, the authors applied two classification algorithms, namely SVM and Stochastic Gradient Descent for classifying subjects into AD, non-AD patients, and Mild Cognitive Impairment (MCI) groups in a cross-visit framework. In \cite{8910399}, the authors extracted a set of features, including the emobase, ComParE \cite{10.1145/2502081.2502224}, eGeMAPS, and MRCG functionals \cite{6905738} and performed three experiments, namely segment level classification, majority vote classification, and active data representation. The authors exploited many classifiers, including Decision Trees, k-Nearest Neighbours, Linear Discriminant Analysis (LDA), Random Forests, and Support Vector Machines. The limitations of these studies lie on the tedious procedure of feature extraction, which demands domain expertise. Also, both studies train shallow machine learning classifiers.

Bertini et al. \cite{BERTINI2022101298}  used the DementiaBank Dataset and employed an autoencoder used in the audio data domain called \textit{auDeep} \cite{JMLR:v18:17-406} and passed the encoded representation (latent vector) to a multilayer perceptron, in order to detect AD patients. Results showed significant improvements over state-of-the-art approaches. The main limitation of this study is the way the speech signal is represented as image. Specifically, the speech signal is converted to a log-Mel spectrogram. On the contrary, the addition of delta and delta-delta features as channels of the image adds more information, since these features add dynamic information to the static cepstral features.

Reference \cite{10.3389/fpsyg.2020.624137}  used the ADReSS Challenge Dataset and exploited i-vectors and x-vectors along with dimensionality reduction techniques for training and testing the shallow machine learning algorithms, namely linear discriminant analysis (LDA), the decision tree (DT) classifier, the k-nearest neighbors classifier, a support vector machine (SVM), and a Random Forest (RF) classifier. The limitation of this study lies on the feature extraction process and the train of traditional machine learning classifiers.

The authors in \cite{10.3389/fcomp.2021.624694} introduced the Open Voice Brain Model (OVBM), which uses 16 biomarkers. Audio files were converted into MFCCs. The ResNet has been used by eight biomarkers for feature representation. Finally, the authors have applied Graph Neural Networks (GNNs) and have extracted a personalized subject saliency map. The limitation of this study lies on the way the speech signal is represented as an image. Specifically, the authors convert the speech signal only to MFCCs. On the contrary, the addition of delta and delta-delta features as channels of the image adds more information, since these features add dynamic information to the static cepstral features. In addition, the authors train multiple models increasing in this way both the training time and computational resources.

\subsection{Multimodal Architectures} \label{multimodal}

Ref. \cite{balagopalan20_interspeech} extracted three sets of features, namely lexicosyntactic, acoustic, and semantic features. In terms of the lexicosyntactic features, the authors extracted the proportion of POS-tags, average sentiment valence of all words in a transcript, and many more. Regarding the acoustic features, MFCCs, fundamental frequency, statistics related to zero-crossing rate, etc. were exploited. With regards to the semantic features, the authors extracted proportions of various information content units used in the picture. Next, they performed feature selection by using the ANOVA and trained four machine learning classifiers, including SVM, neural network, RF, and NB. Results showed that SVM outperformed the other approaches in the multimodal framework. The limitation of this study lies on the way the features from different modalities are combined. More specifically, the authors apply an early fusion strategy, where they fuse the features at the input level. This approach does not capture the inter- and intra-modal interactions. In addition, another limitation is the feature extraction procedure.

The authors in \cite{campbell21_iberspeech} introduced three speech-based systems and two text-based systems for diagnosing dementia from spontaneous speech. Also, they proposed methods for fusing the different modalities. In terms of the speech based systems, the authors extracted i-vectors, x-vectors, and rhythmic features and trained an SVM and a Linear Discriminant Analysis (LDA) classifier. Regarding the text-based models, the authors fine-tuned a BERT model and trained an SVM classifier using linguistic features. Finally, the authors exploited three fusion strategies based on late fusion approach. Therefore, the main limitation of this study is the late fusion approach for fusing the different modalities.

In \cite{9414009}, an early fusion approach was proposed. Specifically, the authors extracted a set of acoustic features, i.e., articulation, prosody, i-vectors, and x-vectors, and a set of linguistic features, including word2vec, BERT, and BERT-Base trained with the Spanish Unannotated Corpora (BETO) embeddings. The authors concatenated these sets of features and trained a Radial Basis Function-Support Vector Machine. The main limitation of this paper is the early-fusion approach.

Research works \cite{chen21r_interspeech} extracted a set of acoustic and linguistic features using the ADReSSo Challenge Dataset. Next, they concatenated these sets of features and trained a Logistic Regression classifier. They also proposed three label fusion approaches,  namely majority voting, average fusion, and weighted average fusion, based on the predictions of several neural networks. The limitations of this study are related to the early and late fusion strategies introduced for detecting AD patients.

Ref. \cite{cummins20_interspeech}  used the ADReSS Challenge Dataset and introduced neural network architectures which use language and acoustic features. Regarding the multimodal approach, the authors fuse the predictions of the three best performing models using a majority vote approach and show that label fusion outperforms the neural networks using either only speech or transcripts. The limitation of this study lies on the usage of a late fusion strategy, i.e., majority vote approach. In this way, multiple models must be trained separately increasing the training time. Also, the inter-modal interactions cannot be captured.

Edwards et al. \cite{edwards20_interspeech}  used the ADReSS Challenge Dataset and exploited both acoustic and linguistic features and combined them in order to detect AD patients. In terms of acoustic features, they applied feature selection techniques including correlation feature selection (CFS) and recursive feature elimination with cross-validation (RFECV), in order to reduce the dimensionality of the respective feature set. Results of the feature selection techniques indicated a surge in the classification performance. The authors chose the ComParE2016 features as the best feature set for further experiments. Regarding linguistic characteristics, they extracted features from transcripts, employed pretrained embeddings (Word2Vec, GloVe, Sent2Vec) or models (ELECTRA, RoBERTa), and exploited phoneme written pronunciation using CMUDict \cite{weide2005carnegie}. The authors stated that the fusion of phonemes and audio features achieved the highest classification results. This study comes with some limitations. Specifically, the authors introduce a feature extraction process and fuse the features of different modalities via an early fusion strategy.

Pompili et al. \cite{pompili20_interspeech}  used the ADReSS Challenge Dataset and extracted also acoustic and linguistic features from speech. Results showed that a simple merge of these feature sets via an early fusion approach did not improve the classification results, since the predictive ability of linguistic features completely overrides acoustic ones. 

Authors in \cite{sarawgi20_interspeech}  used the ADReSS Challenge Dataset and introduced three deep learning architectures for categorizing people into AD patients and non-AD ones, as well as predicting the  Mini-Mental State Exam (MMSE) scores, using disfluency, acoustic, and intervention features. Regarding the multimodal approach, they followed three similar approaches, namely hard, soft, and learnt ensemble. With regards to the hard ensemble, a majority vote was taken between the predictions of the three individual models. Regarding the soft ensemble, a weighted sum of the class probabilities was computed for the final decision. Finally, in terms of the learnt ensemble, a logistic regression voter was trained using class probabilities as inputs. Thus, the limitation of this study is pertinent to the late fusion strategy, where multiple models must be trained and tested separately.

Ref. \cite{pappagari21_interspeech}  used the ADReSSo Challenge Dataset and proposed a similar approach. After training acoustic and language models, the authors explored model fusion by using the output scores of the models as the inputs for a Logistic Regression (LR) classifier to obtain the final prediction. 

Martinc and Pollak \cite{martinc20_interspeech}  used the ADReSS Challenge Dataset and extracted a set of audio, tf-idf, readability, and embedding features and trained traditional machine learning algorithms, including XGBoost, Random Forest (RF), SVM, and Logistic Regression, with Logistic Regression achieving the highest accuracy on the test set. The authors also stated that readability features and the duration of the audio proved to be effective features. The main limitations of this study are: (i) the feature extraction procedure, i.e., demands domain expertise, the best features may not be extracted, (ii) early fusion strategy, and (iii) train of traditional machine learning classifiers.

Koo et al. \cite{koo20_interspeech}  used the ADReSS Challenge Dataset and proposed a deep learning model consisting of BiLSTMs, CNNs, and self-attention mechanism and exploited both textual, i.e., transformer-based models, psycholinguistic, repetitiveness, and lexical complexity features, and acoustic features, i.e., openSMILE and VGGish features. Specifically, they passed each modality through a self attention layer, where key, value, and query corresponded to one single modality. However, the authors concatenated the outputs of the attention layer, which correspond to the two different modalities, and passed them through a CNN layer. The main limitations of this study are pertinent to the feature extraction process and the concatenation of the representation vectors of the two modalities into one vector.

Authors in \cite{pappagari20_interspeech}  used the ADReSS Challenge Dataset and introduced both acoustic and transcript-based models, while they proposed also a method to fuse the two models. More specifically, they first obtained the scores from acoustic and transcript-based models for all utterances from the testing folds during the cross-validation stage and then they employed these predictions in a cross-validation scheme to train and test fusion of scores using a GBR model. The usage of a late fusion strategy constitutes the limitation of this study.

Pan et al. \cite{pan21c_interspeech}  used the ADReSSo Challenge Dataset and introduced also a multimodal approach, where they exploited wav2vec for extracting acoustic embeddings from the audio files and employed also BERT for extracting embeddings for transcripts. Finally, they concatenated these two representations and trained the model for detecting AD patients. The concatenation of the representation vectors of the two different modalities constitutes the main limitation of this study, since the concatenation operation does not capture the inherent correlations between the two modalities.

Zhu et al. \cite{10.3389/fcomp.2021.624683} proposed both unimodal and multimodal approaches. Regarding unimodal models, they employed first MobileNet and YamnNet to discriminate between AD patients and non-AD ones. They converted audio files into MFCC features, duplicated the MFCC feature map twice and made the MFCC feature map as a (p, t, 3)-matrix, in order to match with the module input of the proposed architectures. They used also BERT and Speech BERT. In terms of the multimodal models, the authors exploited Speech BERT, YamnNet, Longformer, and BERT. After extracting the representations of audio and transcripts, they used the add and concatenation operation to fuse these two modalities. Results on the ADReSS Challenge Dataset showed that the concatenation operation of the representations extracted via BERT and Speech BERT outperformed the unimodal models. The limitations of this study are the following: (i) the way the speech signal is represented as an image. More specifically, this study duplicates the MFCC feature map twice and makes the MFCC feature map as a (\textit{p}, \textit{t}, 3)-matrix. On the contrary, the delta and delta-delta features can be used for adding more information \cite{5947425,1168654}. (ii) In terms of the multimodal models, the authors fuse the different modalities via an add and concatenation operation. These methods do not capture the inherent correlations between the two modalities.

Research work \cite{10.3389/fnagi.2021.623607} employed also a bi-modal model consisting of Dense, GRU, CNN, BiLSTM, and attention layers. The authors fused the two modalities by concatenating their respective representations. Results on the ADReSS Challenge Dataset showed an improvement of evaluation results of the multimodal approach over unimodal architectures. The usage of the concatenation operation for fusing the two modalities constitutes a limitation of this study. Also, the feature extraction process proposed by the authors, constitutes another limitation.

Research work \cite{10.3389/fcomp.2021.624659}  used the ADReSS Challenge Dataset and extracted a set of acoustic and linguistic features to train and test algorithms on the AD classification and MMSE prediction task. Regarding the fusion of the two modalities, the authors chose the three best-performing acoustic models along with the best-performing language model, and computed a final prediction by taking a linear weighted combination of the individual model predictions. The method for fusing the two modalities, i.e., late fusion strategy, constitutes the main limitation of this study. Another limitation is the tedious procedure of feature extraction.

In \cite{9459113}, the authors applied majority voting of the acoustic and language models in terms of the multimodal approach. Regarding the acoustic models, they extracted handcrafted features as well as embeddings via deep neural networks, including VGGish, YAMNet, openl3: Music, and openl3: Environment sounds.  In terms of the language models, apart from the handcrafted features, embeddings from BERT base (cased and uncased), BERT large (cased and uncased), RoBERTa (base and fine-tuned on biomedical data), and distilled versions of BERT and RoBERTa, have been exploited. After applying feature aggregation techniques, the authors trained shallow machine learning classifiers for the AD classification and MMSE regression task. The method for fusing the two modalities, i.e., late fusion strategy, constitutes the main limitation of this study.

The authors in \cite{mittal2021multimodal} proposed a late fusion strategy combining both acoustic and text-based models to detect AD patients. Specifically, their text-based model consists of CNNs, BERT, and SentenceBERT, while the proposed acoustic-based model consists of a VGGish model. Finally, the authors combined the two probabilities obtained by the two aforementioned models using a late fusion strategy and obtained an accuracy of 85.30\%. The method for fusing the two modalities, i.e., late fusion strategy, constitutes the main limitation of this study.

In \cite{10.3389/fnagi.2021.642647}, the authors introduced an approach, which accounts for temporal aspects of both linguistic and acoustic features. In terms of the acoustic features, the authors exploited the eGeMAPS feature set, while they used GloVE embeddings with regards to the language features. Next, the Active Data Representation \cite{8910399} with some modifications was employed. The authors used a Random Forest Classifier for performing their experiments. The authors performed a series of experiments and stated that the majority vote approach yielded the best result. The method for fusing the two modalities, i.e., late fusion strategy, constitutes the main limitation of this study.

On the other hand, Rohanian et al. \cite{rohanian20_interspeech} extracted a set of acoustic and linguistic features and employed a neural network architecture consisting of two branches of BiLSTMs, one branch for each modality. In order to control the influence of the two modalities, they introduced feed-forward highway layers \cite{srivastava2015highway} with gating units. Results showed that the multimodal approach yielded better results than the unimodal ones. Research work \cite{rohanian21_interspeech} proposed a similar approach where the authors employed also BERT except for LSTMs to extract embeddings from linguistic features. However, the authors did not experiment with replacing the gating mechanism with a simple concatenation operation. Thus, the addition of the introduced gating mechanism cannot guarantee increase in the performance. Our work differs from \cite{rohanian20_interspeech,rohanian21_interspeech}, since we use BERT and Vision Transformer to extract features from transcripts and image representations of speech respectively. Also, we propose a different multimodal gating mechanism.

\subsection{Related Work Review Findings} \label{overview}

In terms of the architectures using only speech data, it is evident that current research works  \cite{ammar2019evaluation,gauder21_interspeech,balagopalan21_interspeech,10.3389/fpsyg.2020.623237,10.1145/3175587.3175589,8910399,10.3389/fpsyg.2020.624137} have been focused mainly on acoustic feature extraction and then the usage of shallow machine learning algorithms, i.e., SVM, LR, RF etc., or CNNs and BiLSTMs.  The study in \cite{9383491}, which has converted audio files into images of three channels, namely log-Mel spectrograms (and MFCCs), their delta, and delta-delta, has exploited only one pretrained model of the domain of computer vision, i.e., ResNet18.  In addition, the study introduced in \cite{10.3389/fcomp.2021.624683} has converted the audio files into MFCC features, has duplicated the MFCC feature map twice and has made the MFCC feature map as a \textit{(p, t, 3)}-matrix. Next, this study has employed YAMNet, MobileNet, and Speech BERT. However, the limitation of this study lies on the way images are created. On the contrary, delta and delta-delta coefficients are used for recognizing speech better, since the dynamics of the power spectrum, i.e., trajectories of MFCCs over time, are understood better.

Regarding the multimodal models, the majority of the research works have either concatenated or added the representations corresponding to the two different modalities  \cite{10.3389/fcomp.2021.624683,pan21c_interspeech,10.3389/fnagi.2021.623607,koo20_interspeech}. However, the concatenation operation assigns equal importance to each modality and it neglects the inter- and intra-modal interactions. Other research works have trained several language and acoustic models separately and then use majority voting for the final classification of the people as AD patients or non-AD patients  \cite{cummins20_interspeech,sarawgi20_interspeech,9459113,10.3389/fnagi.2021.642647,chen21r_interspeech}. Late fusion approaches have been also proposed including \cite{campbell21_iberspeech,pappagari21_interspeech,pappagari20_interspeech,10.3389/fcomp.2021.624659,mittal2021multimodal,syed20_interspeech,chen21r_interspeech}.  However, these approaches increase substantially the computation time, while the inter-modal interactions are not captured.  In addition, there are studies \cite{9414009,balagopalan20_interspeech,chen21r_interspeech,pompili20_interspeech,martinc20_interspeech,edwards20_interspeech} proposing early fusion approaches, meaning that the features corresponding to the different modalities are concatenated at the input level. None of these works capture the inter- and intra-modal interactions.

Therefore, our work differs significantly from the aforementioned research works, since we \textit{(i)} exploit several pretrained models, including the Vision Transformer, on the vision domain and compare their performances, \textit{(ii)} introduce multimodal models consisting of BERT and Vision Transformer, which can be trained in an end-to-end trainable way, \textit{(iii)} introduce multimodal models with a gating mechanism so as to control the influence of each modality towards the final classification, and \textit{(iv)} introduce the cross-attention mechanism to learn crossmodal interaction and show that crossmodal interaction outperforms the competitive multimodal methods. 

\section{Dataset} \label{dataset}

The ADReSS Challenge dataset \cite{luz20_interspeech} has been used for conducting our experiments. This dataset consists of 78 AD patients and 78 non-AD patients. Each participant (PAR) has been assigned by the interviewer (INV) to describe the Cookie Theft picture from the Boston Diagnostic Aphasia Exam \cite{10.1001/archneur.1994.00540180063015}. This dataset is matched for gender, so as to mitigate bias in the prediction task. Also, it has been carefully selected so as to mitigate common biases often overlooked in evaluations of AD detection methods, including repeated occurrences of speech from the same participant (common in longitudinal datasets), variations in audio quality, and imbalances of age distribution. Moreover, this dataset has been split into a train and test set. Specifically, the train set consists of 108 people, 54 AD and 54 non-AD patients, while the test set comprises 48 people, 24 of whom are AD patients and 24 are non-AD ones.

\section{Proposed Predictive Models using only Speech} \label{unimodal_proposed_models}

In this section, we describe the models used for detecting AD patients using only speech. Our main motivation of exploiting these pretrained models is to find the best performing one and exploit it in the multimodal setting, which will be discussed in detail in Section \ref{proposed_multimodal}. Firstly, we use the Python library librosa \cite{brian_mcfee_2021_4792298} for converting the audio files into Log-Mel spectrograms (and MFCCs), their delta, and delta-delta. We extract Log-Mel spectrograms with 224 Mel bands, window length equal to 2048, hop length equal to 1024, and a Hanning window. For extracting MFCCs, we use 40 MFCCs, a Hanning window, window length equal to 2048, and a hop length of 512. We employ the following pretrained models: \textbf{GoogLeNet (Inception v1)} \cite{7298594}, \textbf{ResNet50} \cite{7780459}, \textbf{WideResNet-50-2} \cite{zagoruyko2016wide}, \textbf{AlexNet} \cite{krizhevsky2014one}, \textbf{SqueezeNet1\_0} \cite{iandola2016squeezenet}, \textbf{DenseNet-201} \cite{8099726}, \textbf{MobileNetV2}\cite{8578572}, \textbf{MnasNet1\_0} \cite{8954198}, \textbf{ResNeXt-50 32$\times$4d} \cite{8100117}, \textbf{VGG16} \cite{simonyan2014very}, \textbf{EfficientNet-B2}\footnote{We experimented with EfficientNet-B0 to B7, but EfficientNet-B2 was the best performing model.} \cite{pmlr-v97-tan19a}, and \textbf{Vision Transformer} \cite{dosovitskiy2021an}.

For all the models, we add a classification layer with two units at the top of the models. Regarding the Vision Transformer, the output of the Vision Transformer ($z_0 ^L$) serving as the image representation is passed through a dense layer with two units in order to get the final output.

\subsection{Experiments}
All experiments are conducted on a single Tesla P100-PCIE-16GB GPU.

\paragraph{\normalsize \textbf{Experimental Setup}} Firstly, we divide the train set provided by the Challenge into a train and a validation set (65-35\%). All models have been trained with an Adam optimizer and a learning rate of 1e-5. We train the proposed architectures five times. We apply \textit{ReduceLROnPlateau}, where we reduce the learning rate by a factor of 0.1, if the validation loss has stopped decreasing for three consecutive epochs. Also, we apply \textit{EarlyStopping} and stop training if the validation loss has stopped decreasing for six consecutive epochs. We minimize the cross-entropy loss function. We test the proposed models using the ADReSS Challenge test set. We average the results obtained by the five repetitions. All models have been created using the PyTorch library \cite{NEURIPS2019_9015}. We have used the Transformers library \cite{wolf-etal-2020-transformers} for exploiting the Vision Transformer\footnote{google/vit-base-patch16-224-in21k}$^{,}$\footnote{We also use the ViTFeatureExtractor.}.

\paragraph{\normalsize \textbf{Evaluation Metrics}} Accuracy, Precision, Recall, F1-Score, and Specificity have been used for evaluating the results of the introduced architectures. These metrics have been computed by regarding the dementia class as the positive one.

\subsection{Results} \label{results_unimodal}

The results of the proposed models mentioned in Section \ref{unimodal_proposed_models}, which receive as input either log-Mel Spectrograms or MFCCs, are reported in Table \ref{compare}. 

In terms of the proposed models with log-Mel Spectrograms as input, as one can easily observe, the Vision Transformer constitutes our best performing model outperforming the other pretrained models in terms of all the evaluation metrics except specificity. To be more precise, Vision Transformer surpasses the other models in accuracy by 2.08-14.58\%, in precision by 1.64-11.22\%, in recall by 5.00-39.17\%, and in F1-score by 2.85-21.85\%. The second best performing model is the AlexNet achieving accuracy and F1-score equal to 62.92\% and 66.91\% respectively. VGG16 constitutes the third best model achieving F1-score and Accuracy equal to 65.55\% and 61.25\% respectively. The other pretrained models achieve almost equal accuracy results ranging from 53.33\% to 59.16\% except for DenseNet-201, which performs very poorly with the accuracy accounting for 50.42\%.

In terms of the proposed models with MFCCs as input, we observe that the Vision Transformer constitutes the best performing model attaining an Accuracy score of 63.33\% and an F1-score of 60.30\%. Specifically, it surpasses the other models in Accuracy by 0.41-9.17\%, in F1-score by 0.10-6.24\%, and in Precision by 0.13-12.85\%. AlexNet is the second best performing model achieving an Accuracy of 62.92\%, while it surpasses the other models in Accuracy by 2.93-8.76\%. MnasNet1\_0, GoogleNet, and VGG16 achieve almost equal accuracy scores ranging from 59.17\% to 59.99\% with the MnasNet1\_0 achieving the highest Accuracy score. Next, SqueezeNet1\_0 and DenseNet-201 yield equal accuracy scores accounting for 58.75\%, with SqueezeNet1\_0 outperforming DenseNet-201 in F1-score by 0.72\%. MobileNetV2 achieves an Accuracy score of 57.92\% followed by EfficientNet-B2, whose accuracy accounts for 57.08. EfficientNet-B2 yields the highest Recall equal to 65.00\%, surpassing the other models by 5.00-10.84\%. ResNeXt-50 32$\times 4d$ achieves the worst accuracy score accounting for 54.16\%. 

In both cases, i.e., log-Mel spectrograms and MFCCs, we observe that Vision Transformer constitutes our best performing model. This can be justified by the fact that all the other pretrained models are based on Convolutional Neural Networks (CNNs). On the contrary, the Vision Transformer does not imply any convolution layer. Specifically, the image is split in patches and is fed to the Vision Transformer network, which exploits the concept of the self-attention mechanism introduced in \cite{NIPS2017_3f5ee243}. Therefore, we believe that the difference in performance is attributable to the transformer encoder, which consists of multi-head self-attention and is implemented in the Vision Transformer.

\begin{longtable}{lccccc}
\caption{Performance comparison among proposed models (using only speech) on the ADReSS Challenge test set. Reported values are mean $\pm$ standard deviation.  Results are averaged across five runs. Best results per evaluation metric and method are in bold.}
\label{compare} \\
\toprule
\multicolumn{1}{l}{}&\multicolumn{5}{c}{\textbf{Evaluation metrics}}\\
\cline{2-6} 
\multicolumn{1}{l}{\textbf{Architecture}}&\textbf{Precision}&\textbf{Recall}&\textbf{F1-score}&\textbf{Accuracy}&\textbf{Specificity}\\ \midrule 
\multicolumn{6}{>{\columncolor[gray]{.8}}l}{\textbf{log-Mel Spectrogram}} \\
\textit{\small{GoogLeNet (Inception v1)}} & 57.01 & 70.00 & 60.92 & 57.08 & 44.17 \\
& $\pm$4.70 & $\pm$19.08 & $\pm$7.43 & $\pm$4.86 & $\pm$24.80 \\ \hline
\textit{\small{ResNet50}} & 58.93 & 41.66 & 47.91 & 55.00 & \textbf{68.33} \\
& $\pm$9.31 & $\pm$6.97 & $\pm$3.61 & $\pm$4.86 & $\pm$14.58 \\ \hline
\textit{\small{WideResNet-50-2}} & 52.99 & 64.16 & 57.70 & 53.75 & 43.33 \\
& $\pm$1.95 & $\pm$10.74 & $\pm$5.39 & $\pm$2.43 & $\pm$8.58 \\ \hline
\textit{\small{AlexNet}} & 60.07 & 75.83 & 66.91 & 62.92 & 50.00 \\
& $\pm$2.60 & $\pm$9.28 & $\pm$5.35 & $\pm$4.04 & $\pm$2.64 \\ \hline
\textit{\small{SqueezeNet1\_0}} & 57.13 & 74.16 & 64.52 & 59.16 & 44.16 \\
& $\pm$2.61 & $\pm$1.66 & $\pm$2.18 & $\pm$3.12 & $\pm$4.99 \\ \hline
\textit{\small{DenseNet-201}} & 50.49 & 70.00 & 58.46 & 50.42 & 30.83 \\
& $\pm$3.58 & $\pm$7.64 & $\pm$3.81 & $\pm$4.82 & $\pm$10.74 \\ \hline
\textit{\small{MobileNetV2}} & 54.92 & 73.33 & 62.69 & 56.66 & 40.00 \\
& $\pm$1.51 & $\pm$7.73 & $\pm$3.70 & $\pm$2.43 & $\pm$4.25 \\ \hline
\textit{\small{MnasNet1\_0}} & 56.66 & 70.00 & 59.84 & 55.83 & 41.66 \\
& $\pm$6.08 & $\pm$22.88 & $\pm$7.42 & $\pm$4.45 & $\pm$28.99 \\ \hline
\textit{\small{ResNeXt-50 32 $\times$ 4d}} & 53.69 & 64.16 & 58.09 & 53.75 & 43.33 \\
& $\pm$3.99 & $\pm$6.24 & $\pm$2.06 & $\pm$4.45 & $\pm$13.59 \\ \hline
\textit{\small{VGG16}} & 58.89 & 74.16 & 65.55 & 61.25 & 48.33 \\
& $\pm$1.18 & $\pm$6.66 & $\pm$3.27 & $\pm$2.12 & $\pm$3.33 \\ \hline
\textit{\small{EfficientNet-B2}} & 54.16 & 58.33 & 55.46 & 53.33 & 48.33 \\
& $\pm$6.44 & $\pm$7.91 & $\pm$3.48 & $\pm$5.53 & $\pm$15.72 \\ \hline
\textit{\small{Vision Transformer (ViT)}} & \textbf{61.71} & \textbf{80.83} & \textbf{69.76} & \textbf{65.00} & 49.16 \\
& $\pm$2.93 & $\pm$6.24 & $\pm$1.61 & $\pm$2.76 & $\pm$10.34 \\ 
\midrule
\multicolumn{6}{>{\columncolor[gray]{.8}}l}{\textbf{MFCCs}} \\
\textit{\small{GoogLeNet (Inception v1)}} & 60.77 & 55.00 & 57.49 & 59.58 & 64.17 \\
& $\pm$3.84 & $\pm$6.66 & $\pm$4.19 & $\pm$3.39 & $\pm$6.77 \\ \hline
\textit{\small{ResNet50}} & 56.38 & 59.16 & 57.30 & 56.25 & 53.33 \\
& $\pm$3.83 & $\pm$8.50 & $\pm$3.45 & $\pm$3.49 & $\pm$12.47 \\ \hline
\textit{\small{WideResNet-50-2}} & 55.87 & 54.16 & 54.06 & 55.00 & 55.83 \\
& $\pm$3.86 & $\pm$11.79 & $\pm$4.51 & $\pm$2.12 & $\pm$14.34 \\ \hline
\textit{\small{AlexNet}} & 65.88 & 55.00 & 59.53 & 62.92 & \textbf{70.83} \\
& $\pm$5.94 & $\pm$7.64 & $\pm$5.13 & $\pm$4.04 & $\pm$8.33 \\ \hline
\textit{\small{SqueezeNet1\_0}} & 58.82 & 59.16 & 58.55 & 58.75 & 58.33 \\
& $\pm$2.89 & $\pm$9.65 & $\pm$5.13 & $\pm$2.76 & $\pm$8.33 \\ \hline
\textit{\small{DenseNet-201}} & 59.40 & 56.66 & 57.83 & 58.75 & 60.83 \\
& $\pm$3.31 & $\pm$4.25 & $\pm$2.22 & $\pm$2.43 & $\pm$6.77 \\ \hline
\textit{\small{MobileNetV2}} & 57.76 & 57.50 & 57.22 & 57.92 & 58.33 \\
& $\pm$2.44 & $\pm$11.61 & $\pm$6.38 & $\pm$3.58 & $\pm$5.89 \\ \hline
\textit{\small{MnasNet1\_0}} & 63.42 & 56.66 & 57.63 & 59.99 & 63.33 \\
& $\pm$10.27 & $\pm$14.81 & $\pm$9.56 & $\pm$6.77 & $\pm$17.95 \\ \hline
\textit{\small{ResNeXt-50 32 $\times$ 4d}} & 53.16 & 60.00 & 55.88 & 54.16 & 48.33 \\
& $\pm$3.19 & $\pm$14.09 & $\pm$8.32 & $\pm$3.73 & $\pm$6.77 \\ \hline
\textit{\small{VGG16}} & 59.20 & 60.00 & 59.49 & 59.17 & 58.33 \\
& $\pm$2.75 & $\pm$3.33 & $\pm$1.61 & $\pm$2.12 & $\pm$5.89 \\ \hline
 \textit{\small{EfficientNet-B2}} & 56.40 & \textbf{65.00} & 60.20 & 57.08 & 49.16 \\
& $\pm$5.89 & $\pm$7.26 & $\pm$5.51 & $\pm$5.98 & $\pm$10.34 \\ \hline
\textit{\small{Vision Transformer (ViT)}} & \textbf{66.01} & 55.83 & \textbf{60.30} & \textbf{63.33} & \textbf{70.83} \\
& $\pm$3.36 & $\pm$4.25 & $\pm$1.89 & $\pm$1.66 & $\pm$5.89 \\
\bottomrule

\end{longtable}

\section{Proposed Predictive Models using Speech and Transcripts} \label{proposed_multimodal}

In this section, we describe the models used for detecting AD patients using transcripts along with their audio files. We have exploited the python library PyLangAcq \cite{lee-et-al-pylangacq:2016} for having access to the manual transcripts, since the dataset has been created using the CHAT \cite{macwhinney2014childes} coding system. For processing the audio files, we use the same procedure mentioned in Section \ref{unimodal_proposed_models}. We mention below the proposed models used in our experiments.

\begin{figure*}[!htb]
\centering
\includegraphics[width=1\textwidth]{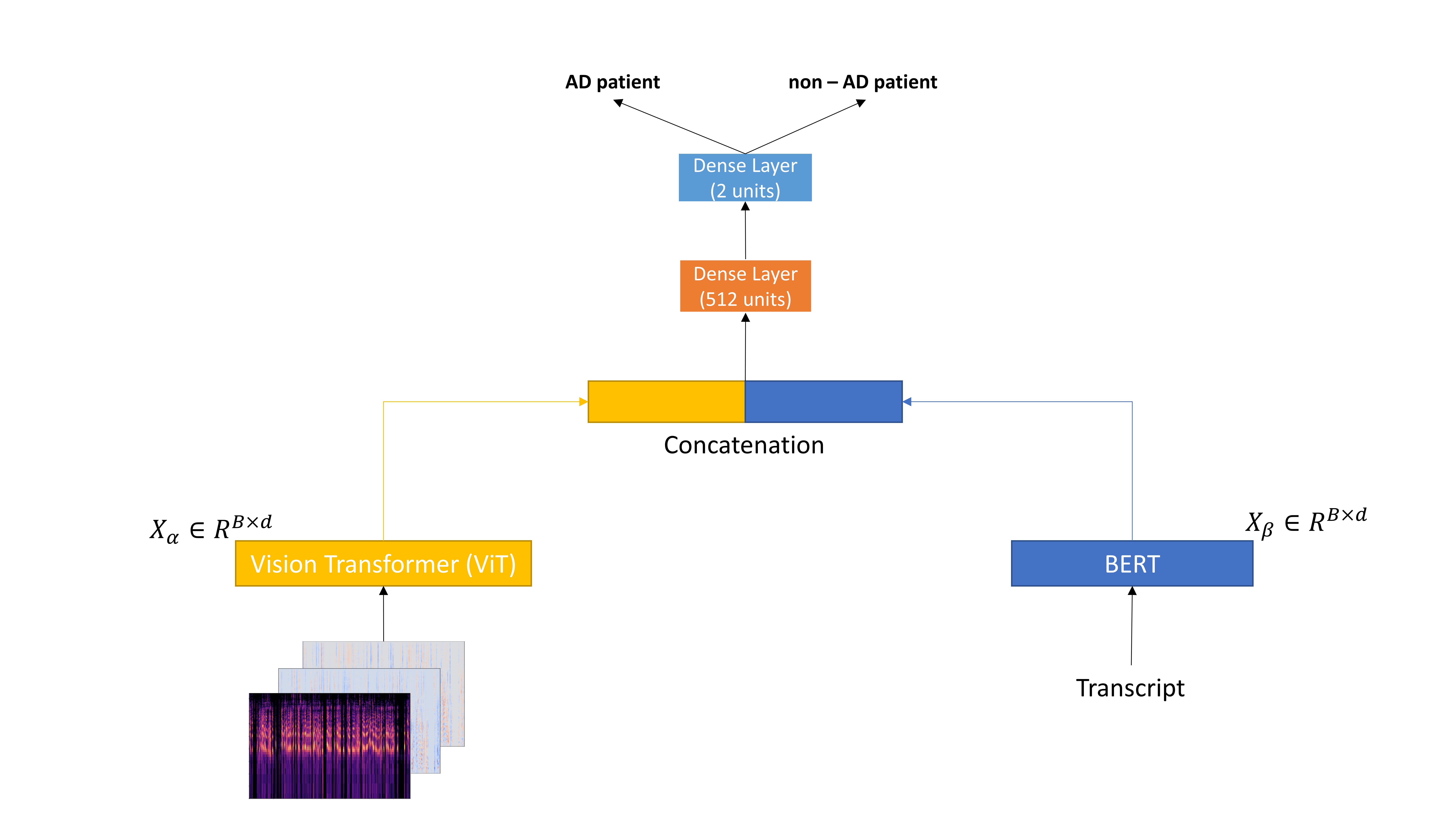}
\caption{BERT + ViT}
\label{bert_vit_concat}
\end{figure*}

\paragraph{\normalsize \textbf{BERT + ViT}} In this model we pass each transcription through a pretrained BERT model \cite{NIPS2017_3f5ee243,devlin-etal-2019-bert} and get the output of the BERT model (CLS token). Regarding the audio files, we convert them into Log-Mel spectrograms  (and MFCCs), their delta, and delta-delta for constructing an image consisting of three channels and pass the image through the ViT. We exploit the Vision Transformer, since it constitutes the best performing model as discussed in Section \ref{results_unimodal}. The output of the ViT ($z_0^L$) is concatenated with the output of the BERT and then the resulting vector is passed through a dense layer with 512 units and a ReLU activation function followed by a dense layer consisting of two units to get the final output. The proposed model is illustrated in Fig. \ref{bert_vit_concat}.

\paragraph{\normalsize \textbf{BERT + ViT + Gated Multimodal Unit}} In this model we pass each transcription through a pretrained BERT model and get the output of the BERT model (CLS token). Regarding the audio files, we convert them into Log-Mel spectrograms  (and MFCCs), their delta, and delta-delta for constructing an image consisting of three channels and pass the image through the ViT. We exploit the Vision Transformer, since it constitutes the best performing model as discussed in Section \ref{results_unimodal}. We get the output of the ViT ($z_0^L$). Next, we employ the Gated Multimodal Unit (GMU) introduced by \cite{arevalo2020gated}, in order to control the contribution of each modality towards the final classification. The equations governing the GMU are described below:

\begin{equation}
    h^t = \tanh{(W^t f^t + b^t)}
\end{equation}
\vspace{-2em}
\begin{equation}
    h^v = \tanh{(W^v f^v + b^v)}
\end{equation}
\vspace{-2em}
\begin{equation}
    z = \sigma(W^z [f^t;f^v] + b^z)
\end{equation}
\vspace{-2em}
\begin{equation}
    h = z * h^t + (1-z)*h^v
\end{equation}
\vspace{-2em}
\begin{equation}
    \Theta	= \{W^t, W^v, W^z\}
\end{equation}

where $f^t$ and $f^v$ denote the text and image representations respectively, $\Theta$ the parameters to be learned, and [.;.] the concatenation operation. Specifically, $W^t \in \mathcal{R}^{128}, W^v \in \mathcal{R}^{128}, W^z \in \mathcal{R}^{128}$. 

The output \textit{h} of the gated multimodal unit is passed through a dense layer consisting of two units.

\begin{figure*}[!htb]
\centering
\includegraphics[width=1\textwidth]{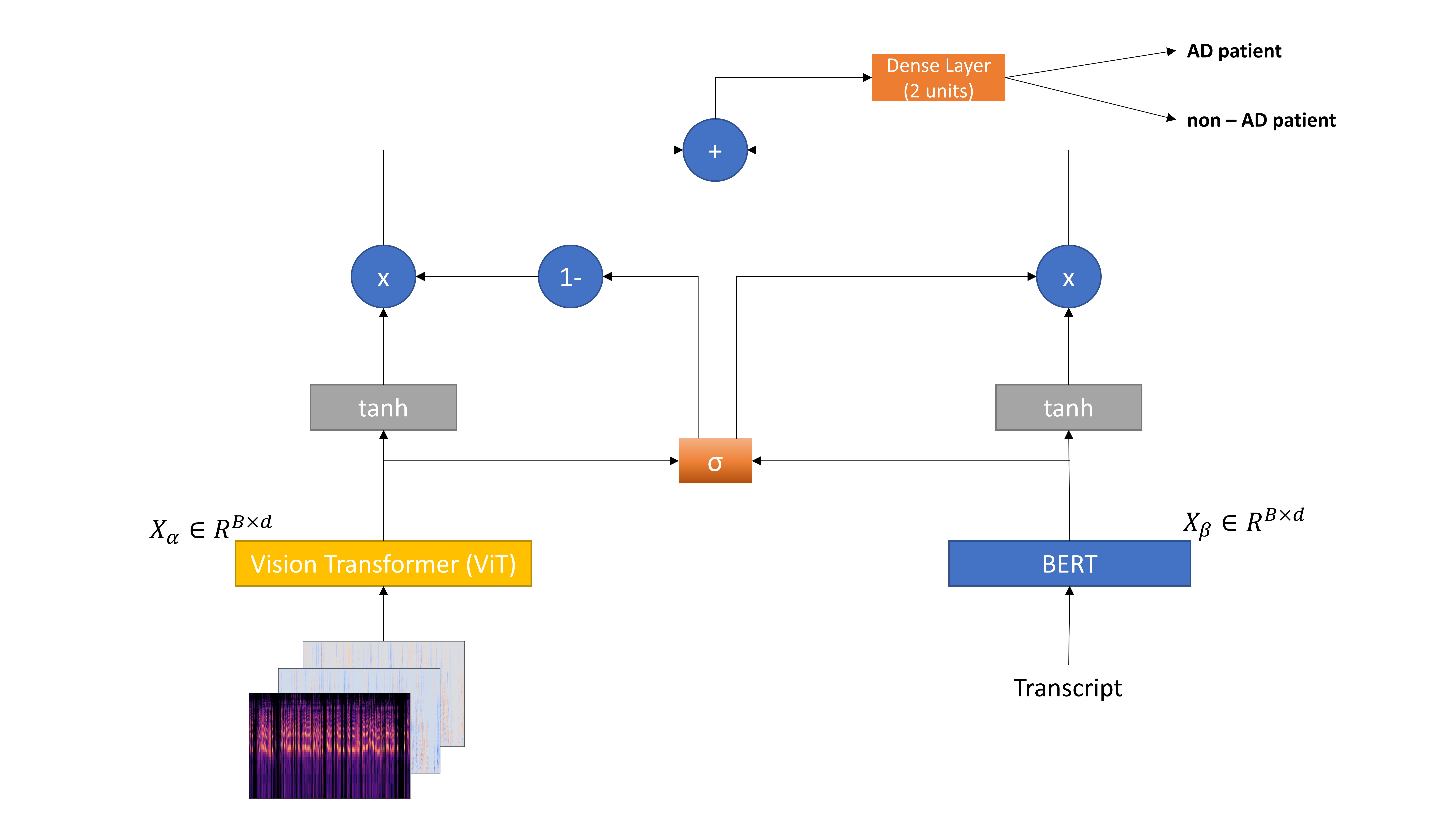}
\caption{BERT + ViT + Gated Multimodal Unit}
\label{bert_vit_gmu}
\end{figure*}

The proposed model is illustrated in Fig. \ref{bert_vit_gmu}.

\paragraph{\normalsize \textbf{BERT + ViT + Crossmodal Attention}} Similar to the previous models, we pass each transcription through a BERT model, and each image through a ViT model. We exploit the Vision Transformer, since it constitutes the best performing model as discussed in Section \ref{results_unimodal}. The image representation can be denoted as $X_\alpha \in \mathbb{R}^{B,T_\alpha,d_\alpha}$, while the text representation can be represented as $X_\beta \in \mathbb{R}^{B,T_\beta,d_\alpha}$, where $B$ constitutes the batch size, $T_{(.)}$ the sequence length, and $d_\alpha$ the feature dimension. Next, we employ the crossmodal attention \cite{tsai-etal-2019-multimodal,sanchez-villegas-aletras-2021-point,9413388}. Specifically, we employ two crossmodal attentions, one from text to image representations and another one from image to text representations. Formally, the crossmodal attention from text to image representation is given by the equations below.

Specifically, we define the queries, keys, and values as:

\begin{equation}
    Q_{\alpha} = X_{\alpha} W_{Q_\alpha}, K_{\beta} = X_{\beta} W_{K_\beta}, V_{\beta} = X_{\beta} W_{V_\beta}
\end{equation}

, where $W_{Q_\alpha} \in \mathcal{R}^{d_\alpha \times d_k}, W_{K_\beta} \in \mathcal{R}^{d_\alpha \times d_k}$, and $W_{V_\beta} \in \mathcal{R}^{d_\alpha \times d_v}$ are learnable parameters. 

Therefore, 

\begin{equation}
    Q_{\alpha} \in \mathcal{R}^{B \times T_{\alpha} \times d_k}, K_{\beta} \in \mathcal{R}^{B \times T_{\beta} \times d_k}, V_{\beta} \in \mathcal{R}^{B \times T_{\beta} \times d_v}
\end{equation}

The latent adaptation from $\beta$ to $\alpha$ is presented as the crossmodal attention, given by the equations below:
\begin{align}
\begin{aligned}
    Y_\alpha &= CM_{\beta \to \alpha} (X_\alpha, X_\beta) \\ &= softmax \left(\frac{Q_\alpha K_\beta ^T}{\sqrt{d_k}} \right) V_\beta \\ &= softmax \left(\frac{X_\alpha W_{Q_\alpha} W_{K_\beta}^T X_\beta ^T}{\sqrt{d_k}}\right) X_\beta W_{V_\beta}
\end{aligned}
\end{align}

The scaled (by $\sqrt{d_k}$) softmax is a score matrix, where the $(i,j)$-th entry measures the attention given by the $i$-th time step of modality $\alpha$ to the $j$-th time step of modality $\beta$. The $i$-th time step of $Y_\alpha$ is a weighted summary of $V_\beta$, with the weight determined by $i$-th row in softmax(·).

Similarly, the crossmodal attention from image to text representation is given by the equations below:

\begin{equation}
    Q_{\beta} = X_{\beta} W_{Q_\beta}, K_{\alpha} = X_{\alpha} W_{K_\alpha}, V_{\alpha} = X_{\alpha} W_{V_\alpha}
\end{equation}

\begin{equation}
    Q_{\beta} \in \mathcal{R}^{B \times T_{\beta} \times d_k}, K_{\alpha} \in \mathcal{R}^{B \times T_{\alpha} \times d_k}, V_{\alpha} \in \mathcal{R}^{B \times T_{\alpha} \times d_v}
\end{equation}

\begin{align}
\begin{aligned}
    Y_\beta &= CM_{\alpha \to \beta} (X_\beta, X_\alpha) \\ &= softmax \left(\frac{Q_\beta K_\alpha ^T}{\sqrt{d_k}} \right) V_\alpha \\ &= softmax \left(\frac{X_\beta W_{Q_\beta} W_{K_\alpha}^T X_\alpha ^T}{\sqrt{d_k}}\right) X_\alpha W_{V_\alpha}
\end{aligned}
\end{align}

\begin{figure*}[!htb]
\centering
\includegraphics[width=1\textwidth]{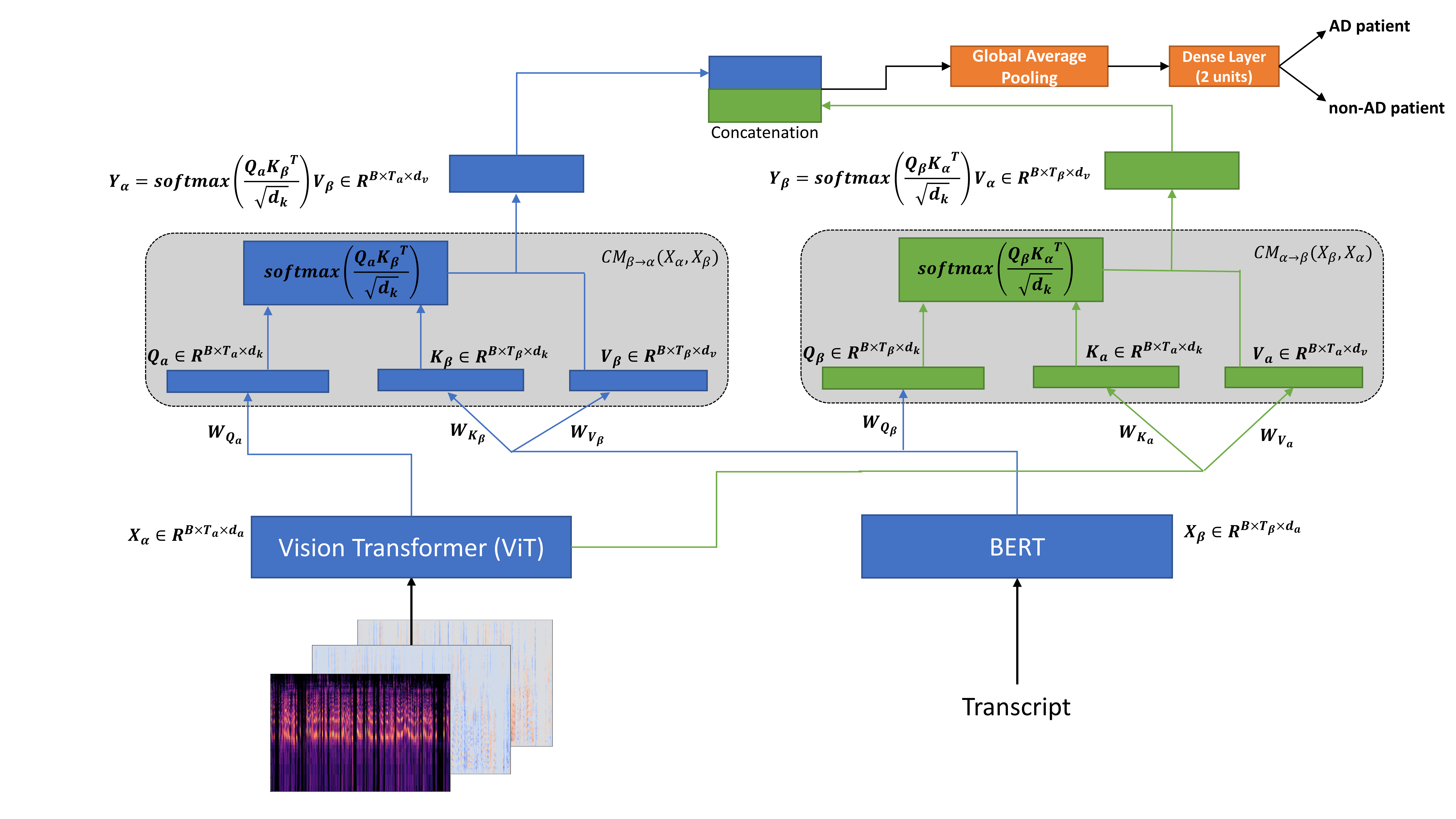}
\caption{BERT + ViT + Crossmodal Attention}
\label{bert_vit_cross}
\end{figure*}

The outputs of the crossmodal attention layers, i.e., $Y_\alpha$ and $Y_\beta$, are concatenated and passed through a global average pooling layer followed by a dense layer with two units. The proposed model is illustrated in Fig. \ref{bert_vit_cross}.

\subsection{Experiments}
All experiments are conducted on a single Tesla P100-PCIE-16GB GPU.

\paragraph{\normalsize \textbf{Comparison with state-of-the-art approaches}}

\begin{enumerate}
\item \textbf{Unimodal state-of-the-art approaches (only transcripts)}
\begin{itemize}
    \item BERT \cite{9769980}: This method trains a BERT model using transcripts.
    \end{itemize}
    \item \textbf{Multimodal state-of-the-art approaches (speech and transcripts)}
    \begin{itemize}
    \item  top-3 late fusion \cite{10.3389/fnagi.2021.642647}: This method proposes a late fusion approach of the three best feature configurations, namely Temporal + char4grams, New + char4grams, and char4grams. The authors train a Random Forest Classifier.
    \item Audio + Text (Fusion) \cite{9459113}: The authors introduce three models for detecting AD patients using only speech data and three models for detecting AD patients using only text data. Finally, they use a majority level approach, where the final prediction corresponds to the class getting the most votes from the six aforementioned models.
    \item  SVM \cite{balagopalan20_interspeech}: This method extracts lexicosyntactic, semantic, and acoustic features, performs feature selection using ANOVA, and finally trains a Support Vector Machine Classifier.
    \item Fusion Maj. (3-best) \cite{cummins20_interspeech}: This method uses a majority vote of three approaches, namely Bag-of-Audio-Words, zero-frequency filtered (ZFF) signals, and BiLSTM-Attention network.
    \item LSTM with Gating (Acoustic + Lexical + Dis) \cite{rohanian20_interspeech}: This research work extracts a set of features from speech and transcripts, passes the respective sets of features through two branches of BiLSTMs, one branch for each modality. Next the authors introduce feed-forward highway layers with a gating mechanism.
    \item System 3: Phonemes and Audio \cite{edwards20_interspeech}: This method transcribes the segment text into phoneme written pronunciation using CMUDict and combines this representation of features with features extracted via the audio.
    \item Fusion of system \cite{pompili20_interspeech}: This method merges features extracted via speech and transcripts and trains a Support Vector Machine Classifier. Features of speech constitute the x-vectors. In terms of the language features, (i) a Global Maximum pooling, (ii) a bidirectional LSTM-RNNs provided with an attention module, and (iii) the second model augmented with part-of-speech (POS) embeddings are trained on the top of a pretrained BERT model.
    \item Bimodal Network (Ensembled Output) \cite{koo20_interspeech}: In this research work, the outputs of the top 5 bimodal networks with high validation results are ensembled and used as the final submission.
    \item GFI, NUW, Duration, Character 4-grams, Suffixes, POS tag, UD \cite{martinc20_interspeech}: This method exploits the gunning fog index, number of unique words, duration of the audio file, character 4-grams, suffixes, pos-tags, and Universal dependency features in a tf-idf setting. Logistic Regression is trained with the corresponding feature sets.
    \item Acoustic \& Transcript \cite{pappagari20_interspeech}: This method employs the scores from the whole training subset to train a final fusion GBR model that is used to perform the fusion of scores coming from the acoustic and transcript-based models for the challenge evaluation.
    \item Dual BERT \cite{10.3389/fcomp.2021.624683}: This method employs a Speech BERT and a Text BERT and concatenates their representations.
    \item Model C \cite{10.3389/fnagi.2021.623607}: This method extracts features from segmented audio and passes them through GRU layers. Regarding the transcripts, this method extracts pos-tags and passes both the transcripts and pos-tags through two separate CNN layers. Then the outputs of the CNN layers are passed through a BiLSTM layer coupled with an Attention Layer. The authors also extract a different set of features from both transcripts and audio files and pass them to a dense layer. The respective outputs are concatenated and passed to a dense layer, which gives the final output.
    \item Majority vote (NLP + Acoustic) \cite{10.3389/fcomp.2021.624659}: This method obtains firstly the best-performing acoustic and language-based models. Next, it computes a weighted majority-vote ensemble meta-algorithm for classification. The authors choose the three best-performing acoustic models along with the best-performing language model, and compute a final prediction by taking a linear weighted combination of the individual model predictions.
\end{itemize}
\end{enumerate}

\paragraph{\normalsize \textbf{Experimental Setup}} 
Firstly, we divide the train set provided by the Challenge into a train and a validation set (65-35\%). Next, we train the proposed architectures five times with an Adam optimizer and a learning rate of 1e-5. We apply \textit{ReduceLROnPlateau}, where we reduce the learning rate by a factor of 0.1, if the validation loss has stopped decreasing for three consecutive epochs. Also, we apply \textit{EarlyStopping} and stop training if the validation loss has stopped decreasing for six consecutive epochs. We minimize the cross-entropy loss function. All models have been created using the PyTorch library \cite{NEURIPS2019_9015}. We use the BERT base uncased version. We test the proposed models using the test set provided by the Challenge. We average the results obtained by the five repetitions.

\paragraph{\normalsize \textbf{Evaluation Metrics}} Accuracy, Precision, Recall, F1-Score, and Specificity have been used for evaluating the results of the introduced architectures. These metrics have been computed by regarding the dementia class as the positive one.

\subsection{\normalsize \textbf{Results}}

The results of the proposed models mentioned in Section \ref{proposed_multimodal} are reported in Table \ref{compare3}. Also this table presents a comparison of our introduced models with both unimodal and multimodal state-of-the-art approaches. 

Regarding our proposed transformer-based models  with log-Mel spectrogram as input, one can observe that BERT+ViT+Crossmodal Attention constitutes our best performing model surpassing the other introduced models in F1-score and Accuracy, while it achieves equal Recall score with BERT+ViT+Gated Multimodal Unit. More specifically, BERT+ViT+Crossmodal Attention outperforms BERT+ViT in recall by a margin of 10.84\%, in F1-score by 3.22\%, and in accuracy by 2.08\%, confirming that the crossmodal attention improves the performance of the multimodal models. Also, it outperforms BERT+ViT+Gated Multimodal Unit in F1-score by 2.77\% and in Accuracy by 3.33\%. In addition, BERT+ViT+Gated Multimodal Unit surpasses BERT+ViT in Recall and F1-score by 10.84\% and 0.45\% respectively. Although BERT+ViT surpasses the other proposed models in Specificity by 6.67-13.34\%, it must be noted that F1-score is a more important metric than Specificity in health-related tasks, since high Specificity and low F1-score means that AD patients are misdiagnosed as non-AD ones.

As one can easily observe, our best performing model, namely BERT+ViT+Crossmodal Attention, surpasses the performance of the multimodal state-of-the-art models, except \cite{10.3389/fnagi.2021.642647,9459113}, in Accuracy by 3.13-15.41\%, while it outperforms the research works in Recall by 3.67-29.17\% and in F1-score by 3.29-18.93\%. At the same time, BERT+ViT+Crossmodal Attention obtains a higher accuracy score than BERT \cite{9769980} outperforming it by 0.83\%. BERT+ViT+Crossmodal Attention outperforms BERT in F1-score by 1.96\%. At the same time, the standard deviations of BERT+ViT+Crossmodal Attention in both F1-score and Accuracy are lower than the standard deviations of BERT \cite{9769980}. This fact indicates the superiority of our introduced model and shows that it can capture effectively the interactions between the two modalities. Regarding BERT+ViT, we can observe that it surpasses the multimodal state-of-the-art models, except \cite{10.3389/fnagi.2021.642647,9459113}, in Accuracy and F1-score by 1.05-13.33\% and 0.07-15.71\% respectively. Thus, the combination of transformer networks, i.e., BERT and ViT, outperforms  or obtains comparable performance to the multimodal state-of-the-art approaches. Although BERT+ViT surpasses Fusion Maj. (3-best) \cite{cummins20_interspeech} in F1-score by a small margin of 0.07\%, it must be noted that our proposed model is more computationally and time effective, since the method in \cite{cummins20_interspeech} trains three different models in order to enhance the classification performance. We observe also that BERT+ViT performs worse than BERT \cite{9769980}. We speculate that this difference of 1.25\% in Accuracy is attributable to the concatenation operation. In terms of BERT+ViT+Gated Multimodal Unit, it also outperforms the state-of-the-art approaches in F1-score and Accuracy except for \cite{cummins20_interspeech,10.3389/fnagi.2021.642647,9459113}. Although BERT \cite{9769980} outperforms BERT+ViT+Gated Multimodal Unit in terms of F1-score and Accuracy, the results show that BERT+ViT+Gated Multimodal Unit can better capture the relevant information of the two modalities on the test set in comparison to the performances of the existing research initiatives proposing multimodal models.

Regarding our proposed transformer-based models with MFCCs as input, one can observe that BERT + ViT + Crossmodal Attention constitutes our best performing model attaining an Accuracy score of 87.92\% and an F1-score of 87.99\%. Specifically, it outperforms the introduced models in Accuracy by 2.50-3.76\%, in F1-score by 1.92-3.65\%, and in Recall by 3.33-10.00\%. Similarly to the proposed transformer-based models with log-Mel spectrogram, we observe that the crossmodal attention yields better results than the concatenation operation and the gated multimodal unit. In addition, we observe that the BERT+ViT+Gated Multimodal Unit surpasses BERT+ViT in Accuracy by 1.26\%. However, BERT+ViT outperforms BERT+ViT+Gated Multimodal Unit in F1-score by 1.73\%.

In comparison with the existing research initiatives, we observe that BERT+ViT+Crossmodal Attention improves the performance obtained by BERT \cite{9769980}. Specifically, Accuracy is improved by 0.42\%, F1-score sees an improvement of 1.26\%, and Recall is improved by 7.50\%. On the contrary, BERT+ViT and BERT+ViT+Gated Multimodal Unit obtain worse performance than BERT \cite{9769980}. Compared with the multimodal state-of-the-art approaches, BERT+ViT+Crossmodal Attention surpasses the research works, except \cite{10.3389/fnagi.2021.642647,9459113}, in Accuracy by 2.72-15.00\%, in F1-score by 2.59-18.23\%, and in Recall by 1.16-26.66\%. BERT+ViT+Gated Multimodal Unit outperforms the research works, except \cite{10.3389/fnagi.2021.642647,9459113}, in Accuracy by 0.22-12.50\%. Finally, BERT+ViT surpasses the research works, except \cite{10.3389/fnagi.2021.642647,9459113,cummins20_interspeech}, in Accuracy by 1.16-11.24\%, while it outperforms the research work \cite{cummins20_interspeech} in F1-score by 0.67\%.

\begin{table*}[!h]
\centering
\caption{Performance comparison among proposed models (using both speech and transcripts) and state-of-the-art approaches on the ADReSS Challenge test set. Reported values are mean $\pm$ standard deviation. Results are averaged across five runs.}
\begin{tabular}{lccccc}
\toprule
\multicolumn{1}{l}{}&\multicolumn{5}{c}{\textbf{Evaluation metrics}}\\
\cline{2-6} 
\multicolumn{1}{l}{\textbf{Architecture}}&\textbf{Precision}&\textbf{Recall}&\textbf{F1-score}&\textbf{Accuracy}&\textbf{Specificity}\\
\midrule
\multicolumn{6}{>{\columncolor[gray]{.8}}l}{\textbf{Unimodal state-of-the-art approaches (only transcripts)}} \\
\textit{BERT \cite{9769980}} & 87.19 & 81.66 & 86.73 & 87.50 & 93.33\\
& $\pm$3.25 & $\pm$5.00 & $\pm$4.53 & $\pm$4.37 & $\pm$5.65 \\
\midrule
\multicolumn{6}{>{\columncolor[gray]{.8}}l}{\textbf{Multimodal state-of-the-art approaches (speech and transcripts)}} \\
\textit{top-3 late fusion \cite{10.3389/fnagi.2021.642647}} & - & - & - & 93.75 & -\\
\hline
\textit{Audio + Text (Fusion) \cite{9459113}} & - & 87.50 & - & 89.58 & 91.67\\
\hline
\textit{SVM \cite{balagopalan20_interspeech}} & 80.00 & 83.00 & 82.00 & 81.30 & 79.00\\
\hline
\textit{Fusion Maj. (3-best) \cite{cummins20_interspeech}} & - & - & 85.40 & 85.20 & -\\
\hline
\textit{LSTM with Gating (Acoustic + Lexical + Dis) \cite{rohanian20_interspeech}} & 81.82 & 75.00 & 78.26 & 79.17 & 83.33\\
 \hline
\textit{System 3: Phonemes and Audio
\cite{edwards20_interspeech}} & 81.82 & 75.00 & 78.26 & 79.17 & 83.33\\
\hline
\textit{Fusion of system \cite{pompili20_interspeech}} & 94.12 & 66.67 & 78.05 & 81.25 & 95.83\\
\hline
\textit{Bimodal Network (Ensembled Output) \cite{koo20_interspeech}} & 89.47 & 70.83 & 79.07 & 81.25 & 91.67\\
\hline
\textit{\makecell[l]{GFI,NUW,Duration,Character 4-grams,Suffixes,\\POS tag,UD \cite{martinc20_interspeech}}} & - & - & - & 77.08 & -\\
\hline
\textit{Acoustic \& Transcript \cite{pappagari20_interspeech}} & 70.00 & 88.00 & 78.00 & 75.00 & 83.00\\
\hline
\textit{Dual BERT \cite{10.3389/fcomp.2021.624683}} & 83.04 & 83.33 & 82.92 & 82.92 & 82.50\\
& $\pm$3.97 & $\pm$5.89 & $\pm$1.86 & $\pm$1.56 & $\pm$5.53\\
\hline
\textit{Model C \cite{10.3389/fnagi.2021.623607}} & 78.94 & 62.50 & 69.76 & 72.92 & 83.33\\
\hline
\textit{Majority vote (NLP + Acoustic) \cite{10.3389/fcomp.2021.624659}} & - & - & - & 83.00 & -\\
\midrule
\multicolumn{6}{>{\columncolor[gray]{.8}}l}{\textbf{Proposed Transformer-based models (log-Mel Spectrogram)}} \\
\textit{\small{BERT+ViT}} & 90.73 & 80.83 & 85.47 & 86.25 & 91.67 \\
& $\pm$2.74 & $\pm$2.04 & $\pm$1.70 & $\pm$1.67 & $\pm$2.64 \\ \hline
\textit{\small{BERT+ViT+Gated Multimodal Unit}} & 80.92 & 91.67 & 85.92 & 85.00 & 78.33 \\
& $\pm$2.30 & $\pm$3.73 & $\pm$2.37 & $\pm$2.43 & $\pm$3.12 \\ \hline
\textit{\small{BERT+ViT+Crossmodal Attention}} & 86.13 & 91.67 & 88.69 & 88.33 & 85.00 \\
& $\pm$3.26 & $\pm$4.56 & $\pm$2.12 & $\pm$2.12 & $\pm$4.25 \\ 
\midrule
\multicolumn{6}{>{\columncolor[gray]{.8}}l}{\textbf{Proposed Transformer-based models (MFCCs)}} \\
\textit{\small{BERT+ViT}} & 86.72 & 85.83 & 86.07 & 84.16 & 86.66 \\
& $\pm$2.05 & $\pm$6.77 & $\pm$2.69 & $\pm$1.02 & $\pm$3.12 \\ \hline
\textit{\small{BERT+ViT+Gated Multimodal Unit}} & 90.57 & 79.16 & 84.34 & 85.42 & 91.66 \\
& $\pm$2.80 & $\pm$5.89 & $\pm$3.53 & $\pm$2.95 & $\pm$2.64 \\ \hline
\textit{\small{BERT+ViT+Crossmodal Attention}} & 87.09 & 89.16 & 87.99 & 87.92 & 86.66 \\
& $\pm$2.40 & $\pm$5.65 & $\pm$2.79 & $\pm$2.43 & $\pm$3.12 \\ 
\bottomrule
\end{tabular}
\label{compare3}
\end{table*}

\section{Discussion}
The identification of dementia from spontaneous speech constitutes a hot topic in recent years due to the fact that it is time and cost-efficient. Although several research works have been proposed towards diagnosing dementia from speech, there are still limitations. For example, most methods extract features from speech or transcripts and train traditional Machine Learning classifiers. Another significant limitation has to do with the way the different modalities, e.g., speech and transcripts, are combined in a single neural network. Specifically, research works train separately speech-based and text-based networks and then use majority voting approaches, thus increasing significantly the training time. Other research works add or concatenate the text and image representations, thus treating equally the two modalities and obtaining suboptimal performance. Furthermore, although transformers have achieved state-of-the-art results in many domains, their potential has not been fully exploited in the task of dementia detection using speech data. To the best of our knowledge, this is the first study employing the Vision Transformer for detecting dementia only from speech. This study aims also to fill gaps with regards to the usage of multimodal models by introducing the Gated Multimodal Unit and the crossmodal attention layers, which have not been applied before in the task of dementia identification from spontaneous speech. From the results obtained in this study, we found that:

\begin{itemize}
    \item \textbf{Finding 1:} The Vision Transformer (receiving as input images consisting of log-Mel spectrogram, delta, and delta-delta) outperformed the other pretrained models, i.e., ResNet50, WideResNet-50-2, AlexNet, etc., in all the evaluation metrics except for Specificity.  Similarly, the Vision Transformer (receiving as input images consisting of MFCCs, delta, and delta-delta) obtained higher scores by the other models in Accuracy, F1-score, and Precision. We believe that the Vision Transformer constitutes our best performing model due to the transformer encoder and the multi-head self-attention. On the contrary, all the other pretrained models are based on convolutional neural networks.
    \item \textbf{Finding 2:} We compared the performance achieved between BERT and BERT+ViT and showed that BERT+ViT achieved slightly worse results. We speculated that this difference may be attributable to the usage of a simple concatenation of the text and image representations. A simple concatenation operation assigns equal importance to the different modalities. In addition, we compared the performance of BERT+ViT on the test set with 13 research works and showed that BERT+ViT outperformed  most of the research works in F1-score and Accuracy. Thus, transformers achieve comparable performance to state-of-the-art approaches.
    \item \textbf{Finding 3:} Results on the ADReSS Challenge test set showed that BERT+ViT+Gated Multimodal Unit (with log-Mel spectrogram) yielded a higher F1-score than BERT+ViT (with log-Mel spectrogram), while BERT+ViT+Gated Multimodal Unit (with MFCCs) yielded a higher Accuracy score than BERT+ViT (with MFCCs). In addition, we compared the performance of BERT+ViT+Gated Multimodal Unit on the test set with 13 multimodal research works and showed that BERT+ViT+Gated Multimodal Unit  achieved comparable performance. 
    \item \textbf{Finding 4:} We presented a new method to detect AD patients consisting of BERT, ViT, and crossmodal attention layers and showed that crossmodal interactions outperform the competitive multimodal models. We compared our best performing model (BERT+ViT+Crossmodal Attention  with log-Mel spectrogram as input) with  13 research works on the ADReSS Challenge test set and showed that our introduced model outperformed  11 of these strong baselines in Accuracy and F1-score by a large margin of 3.13-15.41\% and 3.29-18.93\% respectively. Moreover, the incorporation of the crossmodal attention enhanced the performance obtained by BERT by 0.83\% in Accuracy and by 1.96\% in F1-score. In terms of BERT+ViT+Crossmodal Attention (with MFCCs), we observed that it outperformed  11 of 13 strong baselines in Accuracy and F1-score by a large margin of 2.72-15.00\% and 2.59-18.23\% respectively, while it achieved better performance than BERT. Also, we observed that the variances of BERT + ViT + Crossmodal Attention  by using either log-Mel Spectrogram or MFCCs are lower than BERT \cite{9769980}.
    
    Also, we observed that BERT+ViT+Crossmodal Attention outperforms both BERT+ViT and BERT + ViT + Gated Multimodal Unit. Specifically, BERT+ViT+Crossmodal Attention performs better than BERT+ViT, since BERT+ViT fuses the features of different modalities through a concatenation operation. The concatenation operation ignores inherent correlations between different modalities. In addition, BERT+ViT+Crossmodal Attention outperforms BERT+ViT+Gated Multimodal Unit. This can be justified by the fact that the Gated Multimodal Unit is inspired by the flow control in recurrent architectures, such as GRU or LSTM. Specifically, the Gated Multimodal Unit controls only the information flow from each modality and does not capture interactions between text and image. On the contrary, the usage of the crossmodal attention layers captures the crossmodal interactions, enabling one modality for receiving information from another modality. More specifically, we pass textual information to speech and speech information to text. Therefore, we observe that controlling the flow of information from the two modalities is not sufficient. On the contrary, learning crossmodal interactions is more important.
    
    In addition, we observed that our best performing model, i.e., BERT+ViT+Crossmodal Attention, outperforms most of the strong baselines. This fact justifies our initial hypothesis that early and late fusion strategies and the usage of concatenation or add operation introduced by other studies do not capture effectively the inter-modal interactions of different modalities, thus obtain in this way suboptimal performance.
\end{itemize}

One limitation of the current research work has to do with the limited number of samples in the ADReSS Challenge dataset, i.e., 78 AD and 78 non-AD patients. However, as mentioned in Section \ref{dataset}, one cannot overlook that this dataset is matched for gender and age, so as to mitigate bias in the prediction task. Concurrently, in contrast to other datasets, it has been carefully selected so as to mitigate common biases often overlooked in evaluations of AD detection methods, including repeated occurrences of speech from the same participant and variations in audio quality. Moreover, it is balanced, since it includes 78 AD and 78 non-AD patients. It is also used widely by a lot of research works dealing with the task of dementia identification from speech.

\section{Conclusion and Future Work}

In this paper, we have proposed methods to differentiate AD from non-AD patients using either only speech or both speech and transcripts. Regarding the models using only speech, we exploited several pretrained models used extensively in the computer vision domain, with the Vision Transformer achieving the highest F1-score and accuracy accounting for 69.76\% and 65.00\% respectively. Next, we employed three neural network models in which we combined speech and transcripts. We exploited the Gated Multimodal Unit, in order to control the influence of each modality towards the final classification. In addition, we experimented with crossmodal interactions, where we used the crossmodal attention. Results showed that crossmodal attention can enhance the performance of competitive multimodal approaches and surpass state-of-the-art approaches. More specifically, models incorporating the crossmodal attention yielded accuracy equal to 88.83\% on the ADReSS Challenge test set. In the future, we plan to investigate more methods on how to combine speech and text representations more effectively, including optimal transport. In addition, we aim to use wav2vec2.0 for creating images.


\section*{Declaration of competing interest}

The authors declare that they have no known competing financial interests or personal relationships that could have appeared to influence the work reported in this paper.
\bibliographystyle{cas-model2-names}

\bibliography{cas-refs}



\end{document}